\documentclass[10pt,twocolumn,letterpaper]{article}

\usepackage[pagenumbers]{cvpr} 

\usepackage{silence}
\usepackage[protrusion=true,expansion=true,factor=1100,stretch=30,shrink=30]{microtype}

\renewcommand{\paragraph}[1]{\vspace{.5em}\noindent\textbf{#1.}}

\definecolor{cvprblue}{rgb}{0.21,0.49,0.74}
\usepackage[pagebackref,breaklinks,colorlinks,allcolors=cvprblue]{hyperref}

\usepackage{booktabs}
\usepackage{multirow}

\def\paperID{21}
\def\confName{CVPR}
\def\confYear{2026}

\title{PhysMamba: Selective State Space Models\\
       as Learned Articulated Body Simulators}

\author{Haochuan Zhang \quad Sinisa Todorovic\\
Oregon State University\\
{\tt\small zhanhaoc@oregonstate.edu, sinisa@oregonstate.edu}
}

\begin{document}
\maketitle
\begingroup\renewcommand\thefootnote{}\footnotetext{Accepted (poster) at the PhysHuman Workshop @ CVPR 2026 (non-archival).}\endgroup

\begin{abstract}
We introduce the first learned articulated body simulator based on
a selective state space model (SSM), called PhysMamba. PhysMamba predicts next-frame
full-body state from position, rotation, and joint-action history,
without velocity inputs. We compare four architectures
under partial- and full-observation inputs and three training protocols. The from-scratch rollout training protocol gives Mamba2 strong short- and mid-horizon
accuracy under partial observation (s10\,=\,43~mm, 2/50 diverged), while
the two-stage teacher-based rollout protocol stabilizes GRU but
fails for Mamba2. With CUDA graph compilation, Mamba2 reaches 0.107~ms per frame
(9{,}334~FPS) on an H100 GPU, within 1.1$\times$ of GRU's un-compiled
throughput, adding under 1\% latency to a 30~Hz HMR pipeline and
enabling integration as a differentiable physics module for
video-based mesh recovery.
\end{abstract}
\section{Introduction}
\label{sec:intro}

Recovering 3D human motion from monocular video is 
under-constrained. Physics
priors---gravity, joint limits, contact---could provide additional constraints. For example, analytical-physics
losses~\cite{shimada2020physcap} and engines such as MuJoCo~\cite{todorov2012mujoco}, and differentiable trajectory
optimization~\cite{gartner2022diffphy} apply physics as a post-hoc
regularizer. However, they are
too slow for real-time inference and difficult to train end-to-end. In this paper, for a Human Mesh Recovery  (HMR)
pipeline, we instead seek a fast, differentiable,
state-predictive model, which produces a next-frame state in a
single forward pass, given the current body state and action estimate at the input. Such a model is required to: (i) filter  physically plausible pose candidates per frame, (ii) temporally smooth predictions across the
frames, and (iii) forecast motion to warm-start the HMR
backbone on occlusion-heavy frames. 

We are motivated by recent work on learned neural simulators,  LARP~\cite{andriluka2024larp}, which showed that a 12-layer MLP
with a fixed 20-frame sliding window can predict
next-frame velocities from concatenated state histories, matching
MuJoCo accuracy at ${\sim}10\times$ the throughput. However, LARP cannot adapt its fixed 20-frame temporal context to varying dynamic regimes, while extending the temporal horizon increases the input dimensionality linearly with the number of frames.

Our key observation is that a selective state space model
(SSM) is a natural fit for this problem.
Mamba2~\cite{dao2024mamba2} implements the recurrence
\begin{equation}
  h_{t+1} = \bar{A}_t \cdot h_t + \bar{B}_t \cdot x_t, \quad
  y_t = C_t \cdot h_t + D_t \cdot x_t
\end{equation}
which is structurally analogous to the discrete-time rigid-body
evolution $x_{t+1} = f(x_t, u_t, \mathrm{contact}_t)$, where the $O(1)$
per-step hidden
state $h_t$ represents physical
history. Our contributions include the following:
\begin{enumerate}
  \item We introduce PhysMamba, the first selective SSM-based learned simulator for articulated bodies. Unlike SSM-based mesh regression, which predicts pose from video, PhysMamba predicts future states from current states.
  \item We design a partial-observation training protocol that removes velocity from the input, forcing the model to infer dynamics from position history. Under this setting, we show that our SSM-based temporal integration is competitive with gated and fixed-window baselines at short and medium horizons.
  \item We report a negative result: the standard TF$\to$Rollout strategy stabilizes GRU but fails for Mamba2, causing frequent divergence and extreme gradient growth. We trace this to amplification through the input-dependent $\bar{A}_t$ transition, motivating SSM-specific rollout training as an open problem.
\end{enumerate}
\section{Related Work}
\label{sec:related}

\paragraph{Learned physics simulators}
Graph neural simulators~\cite{allen2023gns,sanchezgonzalez2018gns}
model rigid-body interactions by message passing. For articulated
bodies, SuperTrack~\cite{fussell2021supertrack} uses a fully connected
network to update body-part states, but accumulates joint-displacement
errors over long sequences. LARP~\cite{andriluka2024larp} reduces this
error with displacement-aware features and losses. However, these methods use
memoryless MLPs or fixed windows. Thus, they cannot adapt to dynamics
requiring longer context. We replace the dynamics backbone with a selective SSM, giving adaptive and unbounded temporal memory.

\paragraph{Physics-informed HMR}
PhysCap~\cite{shimada2020physcap} and
DiffPhy~\cite{gartner2022diffphy} add physics through analytic losses
or post-hoc optimization. We instead learn a differentiable 
dynamics model used directly inside the HMR pipeline.

\paragraph{State space models for dynamics}
SSMs are increasingly used for physical dynamics. Phy-SSM
~\cite{wang2025physsm} splits the transition matrix into known-physics
and learned-residual terms for long-horizon vehicle and drone
forecasting. However,
prior SSM work does not address articulated rigid bodies with contact.
This setting is harder: it has high-dimensional states
(24 bodies $\times$ position/rotation/velocity), discontinuous contact
changes, and kinematic-chain constraints. We adapt selective SSMs to
this domain and analyze when adaptive temporal memory helps.
\section{Our Approach}
\label{sec:method}

\begin{figure}[t]
\centering
\includegraphics[width=\linewidth]{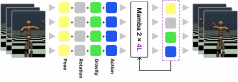}
\caption{PhysMamba overview. Per-frame inputs (pose, rotation, gravity,
action) are concatenated into a 285d vector and fed sequentially into a
4-layer Mamba2 SSM. The recurrent hidden state accumulates temporal
context; the output head predicts the 358d next-frame physical state.}
\label{fig:arch}
\end{figure}

We formulate learned articulated body simulation as a
sequence-to-sequence prediction problem. Given a sequence of physical
states and control actions, the objective is to predict the next-frame
state:
\begin{equation}
  \hat{s}_{t+1} = f_\theta(s_{\leq t},\, a_{\leq t})
\end{equation}
where $s_t \in \mathbb{R}^{358}$ is the full physical state of a
24-body humanoid at time $t$, $a_t \in \mathbb{R}^{69}$ is the control
action (RL policy output), and $f_\theta$ is the learned dynamics
model.

\paragraph{State representation}
Following PHC~\cite{luo2023phc}, the state $s_t$ is a 358-dimensional
vector encoding the humanoid's proprioception in a heading-aligned,
root-relative coordinate frame:
\begin{equation}
  s_t = [h_{\mathrm{root}},\; p_{\mathrm{body}},\;
         R_{\mathrm{body}},\; v_{\mathrm{body}},\;
         \omega_{\mathrm{body}}]
\end{equation}
where $h_{\mathrm{root}}$ is the root height,
$p_{\mathrm{body}} \in \mathbb{R}^{23 \times 3}$ are root-relative
body positions, $R_{\mathrm{body}} \in \mathbb{R}^{24 \times 6}$ are
6D rotation representations~\cite{zhou2019continuity}, and
$v_{\mathrm{body}}, \omega_{\mathrm{body}} \in \mathbb{R}^{24 \times
3}$ are linear and angular velocities.

\paragraph{Input modes}
We study two input configurations:
\begin{itemize}
  \item \textbf{Full-state (430d):} The model receives the complete
    Markov state, $x_t = [s_t, g, a_t]$ where
    $g = [0, 0, {-9.81}]$ is gravity. 
  \item \textbf{Partial-observation (285d):}
    $x_t = [p_{\mathrm{body}}, R_{\mathrm{body}}, g, a_t]$. Velocities
    $(v, \omega)$ are withheld. The model must infer velocity from
    position/rotation history via temporal integration---a very challenging setting
    for memoryless models.
\end{itemize}

\subsection{PhysMamba Architecture}

PhysMamba uses stock Mamba2~\cite{dao2024mamba2} blocks in a pre-norm
residual stack:
\begin{equation}
  h^{(l)} \leftarrow h^{(l)} +
  \mathrm{Mamba2}\!\left(\mathrm{RMSNorm}(h^{(l)})\right)
\end{equation}
with an input projection from $\mathbb{R}^{D_{\mathrm{in}}}$ to
$\mathbb{R}^{d_{\mathrm{model}}}$ and an output head to
$\mathbb{R}^{358}$. The Mamba2 selective scan provides input-dependent state transitions:
\begin{equation}
  h_t = \bar{A}_t h_{t-1} + \bar{B}_t x_t
\end{equation}
where $\bar{A}_t, \bar{B}_t$ are discretized from continuous
parameters using the Structured State-space Duality (SSD) framework. Both are input-dependent.

\begin{table}[ht]
\centering\small
\begin{tabular}{lc}
\toprule
Hyperparameter & Value \\
\midrule
$d_{\mathrm{model}}$ & 512 \\
$d_{\mathrm{state}}$ & 64 \\
head\_dim & 64 \\
$n_{\mathrm{layers}}$ & 4 \\
expand & 2 \\
Parameters & $\sim$7M \\
\bottomrule
\end{tabular}
\caption{PhysMamba architecture configuration.}
\label{tab:arch}
\end{table}

\paragraph{Baselines}
All baselines predict a 358d next-state from the input:
\textbf{GRU} ($d{=}680$, 2 layers, $\sim$6M);
\textbf{MLP} ($d{=}500$, 12 layers, ELU, $\sim$3M, no temporal
context);
\textbf{MLP-LARP} ($d{=}440$, 12 layers, ELU, window${=}20$,
$\sim$6M, autoregressive rollout training with gradient batch-drop at
norm $> 0.3$).

\subsection{Training Data}
\label{sec:data}

We generate training data using the Perpetual Humanoid Controller
(PHC)~\cite{luo2023phc}, a reinforcement learning policy that tracks
motion capture sequences in MuJoCo simulation:
\[
  \text{AMASS (CMU, 2088 clips)}
  \;\xrightarrow{\text{PHC inference}}\;
  \{(s_t, a_t, s_{t+1})\}
\]
We roll out PHC on 2088 AMASS CMU clips; 40 are dropped due to PHC
tracking failure or clip length below the minimum rollout horizon,
yielding 2048 trajectories (1847 train / 201 val, $\sim$785k frames
at 30~Hz).
Each transition is the result of actual physics simulation --- the next
state $s_{t+1}$ is computed by MuJoCo given the current state and the
PHC policy's action, not kinematic interpolation. Per-batch z-axis
rotation augmentation is applied following LARP.

\subsection{Loss Function}

For Mamba2 / GRU / MLP we use the following loss:
\begin{equation}
  \mathcal{L} = \mathcal{L}_p + \mathcal{L}_r
               + 0.1\,\mathcal{L}_v + 0.1\,\mathcal{L}_d,
\end{equation}
where $\mathcal{L}_p$ is position MSE over 23 body joints;
$\mathcal{L}_r = 1 - |q_{\mathrm{pred}} \cdot q_{\mathrm{gt}}|$ is
quaternion rotation distance averaged over 24 bodies (bounded in
$[0,1]$ to prevent gradient spikes through SSM recurrence);
$\mathcal{L}_v$ is velocity MSE over $v_{\mathrm{body}}$ and
$\omega_{\mathrm{body}}$ ($24{\times}3 + 24{\times}3 = 144$d); and $\mathcal{L}_d$ is bone
displacement MSE (LARP Eq.~4) to enforce kinematic chain consistency.
MLP-LARP omits $\mathcal{L}_v$ following LARP's original design.

\subsection{Two-Stage Training}

Training learned simulators faces a fundamental challenge: teacher
forcing (TF) provides clean gradients for accurate single-step prediction, but the resulting models can collapse during autoregressive rollout due to distribution shift. Pure rollout training addresses
stability but converges slowly and sacrifices single-step accuracy.
We address this with a two-stage strategy inspired by the
pretrain-then-distill paradigm in physics-based motion representation
learning~\cite{luo2024pulse}:

\paragraph{Stage 1: Teacher-forced pre-training}
Train $f_\theta$ on ground-truth input sequences with standard
backpropagation through full clips. The model learns accurate
state-to-state mappings with clean gradients.

\paragraph{Stage 2: Autoregressive rollout fine-tuning}
Initialize from the Stage 1 checkpoint. For each training clip, warm
up the hidden state on $K \sim \mathrm{Uniform}(5,50)$ ground-truth
frames (no gradient), then unroll for
$N_h \sim \mathrm{Uniform}(20,150)$ steps (Mamba2) or
$\mathrm{Uniform}(20,100)$ steps (GRU), feeding the model's own
predictions. Gradients are truncated at TBPTT chunks of 50 (Mamba2)
or 20 (GRU) steps:
\begin{equation}
  \hat{s}_{t+1},\, h_{t+1}
  = f_\theta\!\left(\mathrm{build\_input}(\hat{s}_t, a_t^{\mathrm{gt}}),\, h_t\right).
\end{equation}
The loss is accumulated over the rollout horizon and backpropagated
through the recurrent chain. GRU Stage~2 uses a reduced learning rate
($10^{-4}$ vs.\ $3{\times}10^{-4}$); Mamba2 inherits the Stage~1
rate ($3{\times}10^{-4}$), as  lowering
the learning rate accelerated rather than prevented divergence.
\section{Experiments}
\label{sec:exp}

\paragraph{Data}
AMASS~\cite{mahmood2019amass} CMU subset $\to$ PHC inference $\to$
1847 train / 201 val clips, $\sim$785k frames at 30~Hz.

\paragraph{Training}
AdamW, cosine schedule, batch 8 clips, z-rotation augmentation.
Stage~1 uses lr $3{\times}10^{-4}$; Stage~2 uses $10^{-4}$ for GRU
and $3{\times}10^{-4}$ for Mamba2 (see \S\ref{sec:method}).

\paragraph{Evaluation}
(1)~\textit{Single-step MPJPE}: teacher forcing on val set.
(2)~\textit{Rollout}: 150-step autoregressive rollout from GT frame~0;
divergence = any step $>500$~mm (50 val clips).
(3)~\textit{Physical-validity addendum} (per reviewer concern \#3):
foot-slide in mm/frame on 2-frame GT-contact pairs
(SimPoE~\cite{yuan2021simpoe} joint-level proxy) and
PhysCap~\cite{shimada2020physcap} mean-penetration-error MPE (pen
depth conditioned on GT in-contact frames). To make these metrics
meaningful, we retrain the partial-obs Mamba2 / GRU / MLP-LARP
rollout models with a root-height channel and a contact-conditioned
$L_{fs}$ term added to the v1 loss; the retrain setup, full metric
definitions, and numbers are in supplementary~\S S1--S5.

\subsection{Main Results}

\begin{table}[ht]
\centering\small
\begin{tabular}{lcc}
\toprule
Model & 285d (partial-obs) & 430d (full-state) \\
\midrule
Mamba2    & \textbf{29.2} & 37.5 \\
GRU       & 36.6 & \textbf{29.2} \\
MLP       & 49.4 & 49.9 \\
MLP-LARP  & 88.3 & 86.0 \\
\bottomrule
\end{tabular}
\caption{Single-step prediction (teacher forcing, val MPJPE in mm).
Bold = best per column.}
\label{tab:singlestep}
\end{table}

\begin{table}[ht]
\centering\small
\setlength{\tabcolsep}{3pt}
\begin{tabular}{llccccc}
\toprule
Model & Training & s1 & s10 & s50 & s150 & Div. \\
\midrule
Mamba2 285d & TF          & 23.0 & 692 & 808 & 862 & 50/50 \\
GRU 285d    & TF          & 27.9 & 514 & 584 & 594 & 49/50 \\
MLP-LARP 430d & Rollout   & 60.3 & 110 & 246 & 482 & 47/50 \\
Mamba2 285d & Rollout     & 43.6 & \textbf{43.1} & 91.7 & 184  & 2/50  \\
GRU 285d    & TF$\to$Ro   & 45.6 & 56.4 & 114.7 & \textbf{147} & 2/50 \\
Mamba2 285d & TF$\to$Ro$^\dagger$ & 105 & 174 & 203 & 242 & 33/50 \\
\bottomrule
\end{tabular}
\caption{Autoregressive rollout (150 steps, MPJPE in mm, 285d input unless
noted). ``Div.''~counts clips where any step exceeds 500~mm (50 val clips).
$^\dagger$Gradient instability, see \S\ref{sec:analysis}.}
\label{tab:rollout}
\end{table}

\begin{figure}[ht]
\centering
\includegraphics[width=\linewidth]{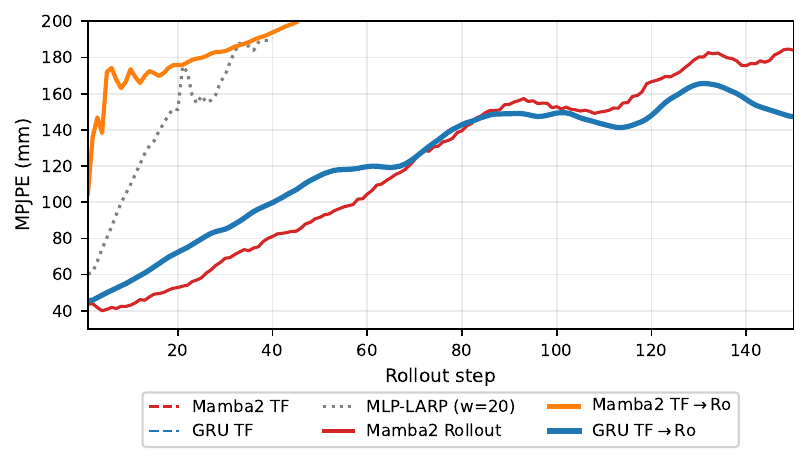}
\caption{Per-step MPJPE over 150-step autoregressive rollout (GT actions,
50 clips). All models use 285d partial-obs input except MLP-LARP (430d
full-state). All TF-only models diverge within 10 steps ($>$500~mm,
not shown). Rollout-trained Mamba2 and two-stage GRU TF$\to$Ro both
remain stable across 150 steps.}
\label{fig:rollout}
\end{figure}

\subsection{Analysis}
\label{sec:analysis}

\paragraph{Partial observation reveals SSM advantage}
The 285d/430d comparison reveals a clean \emph{architecture flip}.
Under 430d (velocity available), GRU achieves \textbf{29.2~mm}
vs.\ Mamba2's 37.5~mm---the Markov state suffices. Under 285d
(velocity withheld), the roles reverse: Mamba2 achieves
\textbf{29.2~mm} vs.\ GRU's 36.6~mm ($+$20\%).
Each architecture reaches the \emph{same peak accuracy} (29.2~mm)
only in its preferred regime; MLP shows no flip (49.4 vs.\ 49.9~mm),
confirming the gain comes from temporal integration.

\paragraph{Accuracy vs.\ stability tradeoff}
All TF-trained models diverge within steps 3--10, consistent with
distribution shift. Among rollout-trained models, Mamba2 Rollout
achieves the best accuracy (s10\,=\,\textbf{43.1}~mm, 2/50 diverged),
while GRU TF$\to$Ro achieves the lowest long-horizon error
(s150\,=\,\textbf{147}~mm, 2/50 diverged). MLP-LARP rollout reduces
but does not eliminate divergence (47/50).

\paragraph{Two-stage recipe fails for Mamba2}
GRU TF$\to$Ro succeeds (2/50 diverged, s150\,=\,147~mm), but
Mamba2 TF$\to$Ro diverges in 33/50 clips with gradient norm reaching
$|\mathbf{g}|{>}10^{16}$ during Stage~2. We hypothesize this stems
from Mamba2's input-dependent $\bar{A}_t$ creating a non-stationary
recurrence whose Jacobian amplifies gradient noise under BPTT, whereas
GRU's fixed gating is better-conditioned.
We attempted multiple mitigations---reduced learning rate
($5{\times}10^{-5}$), fixed-rollout curriculum, and longer warmup
phases---none prevented divergence. Gradient norms still
exceeded $10^{16}$ within 30 epochs, suggesting the instability is
structural rather than hyperparameter-sensitive. SSM-specific rollout
training strategies (\eg, log-barrier on spectral radius of
$\bar{A}_t$, state-space Jacobian clipping, or forward-mode state
gradient estimation) remain an open direction.

\paragraph{Physical validity}
On the v2 retrain (286d + $L_{fs}$, supp.~\S S5), Mamba2 attains the
lowest foot-slide and the lowest PhysCap MPE among the three
matched-recipe configurations; the contact-conditioned $L_{fs}$
rollout loss supervised by the MuJoCo body-contact mask is what
drives this gap.

\paragraph{Latency}
Per-step inference (batch\,=\,1, H100): Mamba2 1.56~ms (641~FPS), GRU
0.097~ms (10{,}352~FPS), MLP ${\sim}$0.21~ms. The apparent 16$\times$
gap stems from Python dispatch overhead in Mamba2's selective scan
rather than from the scan kernel itself: with CUDA graph compilation,
Mamba2 reaches 0.107~ms (9{,}334~FPS)---within 1.1$\times$ of GRU's
un-compiled throughput. GRU's cuDNN-fused kernel already minimizes
dispatch, so CUDA graph would yield diminishing returns; a like-for-like
compiled comparison is left to future work. Both models add under 1\%
latency to a 30~Hz HMR pipeline.
\section{Conclusion}
\label{sec:conclusion}

We presented \textbf{PhysMamba}, a selective SSM for learning
articulated body dynamics, and reported three findings as follows: (1)~under partial observation, the SSM hidden state
integrates physical history more flexibly than memoryless and
fixed-window baselines at single-step prediction, where Mamba2
attains the best 285d MPJPE among the four architectures we
evaluated; (2)~training protocol is as load-bearing as
architecture---a single Mamba2 backbone spans a range of accuracy
and stability outcomes depending on the recipe, and at long
horizons GRU TF$\to$Ro is the lowest-error configuration; (3)~the same input-dependent $\bar{A}_t$ that powers
Mamba2's forward pass resists BPTT stabilisation---three
mitigations failed to prevent divergence, leaving SSM-specific
rollout training (spectral regularization of $\bar{A}_t$,
forward-mode gradients) an open problem~\cite{wang2025physsm}.

\paragraph{Long-horizon hidden-state behavior as a Mamba-family
issue}
The GRU long-horizon win in our table is consistent with a broader
pattern in the selective-SSM literature: while Mamba's
input-dependent recurrence is highly expressive on short and
medium contexts, controlling the hidden state's behaviour over
\emph{long} unrolled horizons remains structurally harder than for
gated RNNs. The forget gate of a GRU is bounded and per-dimension
contractive by construction, which produces a stable long-context
prior; Mamba's per-step transition matrix is unbounded and
input-dependent, so without an explicit constraint nothing prevents
the effective recurrence radius from drifting over long rollouts.
We therefore see SSM-based simulators as needing a separate class of
\emph{robust hidden-state controllers}---e.g.\ spectral or Lyapunov
constraints on $\bar{A}_t$, forward-mode state-gradient estimators,
or learned dissipation terms---rather than direct application of
the GRU recipe. We treat this as the principal direction beyond this
paper.

\paragraph{Toward physics-aware HMR}
Our target application is video-based human mesh recovery, where
PhysMamba serves as a fast differentiable dynamics module. Given
per-frame pose candidates $\{\hat{s}_t\}$ from an HMR backbone, the
one-step prediction $\hat{s}_{t+1} = f_\theta(\hat{s}_t, \hat{a}_t)$
provides a physically consistent expectation to regularize against,
turning per-frame noise into smooth trajectories. Rolling the hidden
state $h_t$ forward 5--15 frames yields a \emph{motion intent
forecast}---a warm start for the HMR backbone on occluded frames,
and an anticipatory signal for downstream tasks (e.g., biomechanics, AR
avatars) where reacting to observed pose may be
too late.

\paragraph{Limitations}
Rollout evaluation uses ground-truth PHC actions rather than a
closed-loop policy, isolating simulator accuracy from controller
error but leaving the end-to-end HMR loop unverified. Integrating
PhysMamba behind a published HMR controller and reporting downstream
reconstruction quality is beyond the scope of this paper. Evaluation is on a single
humanoid morphology (SMPL 24-body) and a single dataset (AMASS-CMU). 
Cross-morphology and cross-dataset transfer are not assessed.
Contact is supervised implicitly from MuJoCo body-contact labels
rather than modeled at the mesh-vertex level. The Mamba2 BPTT
instability is characterized, not solved.

\section*{Supplementary Material --- PhysMamba}

\noindent\textbf{Note.} This supplementary collects definitions and
secondary results that did not fit in the 4-page main paper.

\subsection*{S1 --- Foot-slide metric: definition and caveat}
\label{sec:supp_footslide}

The foot-slide number referenced in the main paper's Physical-validity
addendum (\S4.3) is a \emph{joint-level proxy} of the foot-skate metric
defined by SimPoE~\cite{yuan2021simpoe}, which uses SMPL mesh vertices.
The procedure is:
\begin{enumerate}
  \item Identify the 4 foot joints in the PHC DFS body order:
    L\_Ankle (idx~2), L\_Toe (idx~3), R\_Ankle (idx~6), R\_Toe (idx~7).
  \item Compute the world-frame Z of each foot joint via
    $z_{\mathrm{world}} = \mathrm{root\_h} + \mathrm{body\_pos}_z$.
  \item For each joint $k$ and consecutive timestep pair $(t, t{+}1)$,
    require \emph{both} $z^{\mathrm{GT}}_k(t) < 5\,\mathrm{cm}$ and
    $z^{\mathrm{GT}}_k(t{+}1) < 5\,\mathrm{cm}$ (the joint is on the
    ground in two adjacent frames; this is the 2-frame contact
    criterion).
  \item On qualifying $(k, t)$ pairs, sum the predicted joint's XY
    displacement
    $\| \hat{p}_k(t{+}1) - \hat{p}_k(t) \|_2$.
  \item Report mean over qualifying pairs in mm/frame.
\end{enumerate}

The SimPoE original uses SMPL mesh vertices that contact the ground,
not joint centres. As a result our number is in mm/frame but is not
directly comparable to SimPoE's value of 3.4~mm/frame on H3.6M; we
expect a joint-centre version to under-report slide for stationary
contact (where the foot mesh deforms but the centre stays fixed) and
over-report it for fast contact transitions. A future revision will
upgrade to mesh-vertex evaluation using SMPL forward kinematics.

\subsection*{S2 --- PhysCap MPE: definition}

We report the mean penetration error (MPE) following
PhysCap~\cite{shimada2020physcap}:
\begin{equation*}
  \mathrm{MPE} = \frac{1}{|\mathcal{C}_{\mathrm{GT}}|}
    \sum_{(t,k) \in \mathcal{C}_{\mathrm{GT}}}
    \max(0,\; -\hat{z}^{\mathrm{toe}}_k(t))
\end{equation*}
where the GT-in-contact set is
$\mathcal{C}_{\mathrm{GT}} = \{(t,k) : c_t[k] = 1,\, k \in \{3,7\}\}$
and $\hat{z}^{\mathrm{toe}}_k(t)$ is the predicted toe-joint world-frame
height. Unlike the unconditioned penetration frame rate, MPE counts
pen depth \emph{only on frames where the GT toe is in contact with
the floor}, which is the regime where the physical floor constraint
is active.

\subsection*{S3 --- Contact-mask source}

The main paper $L_{fs}$ uses the MuJoCo body-contact indicator
exported at data-generation time (\S\ref{sec:data}). Each frame
carries a 23-d binary vector $c_t$; for $L_{fs}$ we read entries 3
and 7 (L\_Toe, R\_Toe). AMASS itself carries no contact annotation;
all contact labels in this work come from the MuJoCo rollout.

\subsection*{S4 --- Response to OpenReview reviewers}

\paragraph{Concern \#1: HMR end-to-end (TWiz, vUfK).}
All rollouts in this paper use GT PHC actions; the simulator is not
shown behind a real HMR backbone. A full HMR substitution requires
re-running PHC inference with a published HMR controller as the
action source, which exceeds the camera-ready window of a
non-archival workshop. The body-part perturbation table in~\S S6
partially addresses the underlying robustness question (how
sensitive each architecture is to controller noise). The extended
journal version will close the HMR loop.

\paragraph{Concern \#2: Long-horizon stability (TWiz, vUfK).}
Mamba2 Rollout's strength is short- and mid-horizon
(s1\,=\,43.6~mm, s10\,=\,\textbf{43.1}~mm; Table~\ref{tab:rollout}),
not long-horizon: GRU TF$\to$Ro is in fact the lowest-error row at
s150 (\textbf{147}~mm vs.\ Mamba2 Rollout's 184~mm), and we now state
this explicitly in the abstract and intro. Long-horizon stability
under selective-SSM BPTT is the open problem we report as a
negative result.

\paragraph{Concern \#3: Physical validity (7hya).}
Addressed by the physical-validity addendum: foot-slide and PhysCap
MPE measured on a v2 retrain of the partial-obs rollout models that
adds a root-height channel ($h_{\mathrm{root}}$, 286d input) and a
contact-conditioned $L_{fs}$ term supervised by the MuJoCo body
contact mask (\S S3 above). Definitions: foot-slide
(\S\ref{sec:supp_footslide}), PhysCap MPE (\S S2). The retrain
configuration and numbers are in~\S S5.

\paragraph{Concern \#4: Action / data-generation policy (7hya).}
Section~\S S6 reports per-body-group sensitivity of the v1
rollout-trained checkpoints to action perturbations. PHC tracking
quality on the kept AMASS-CMU clips is documented in~\cite{luo2023phc}.
The 40 clips removed from v1 were dropped because the PHC controller
itself diverged before the end of the source AMASS sequence
(manipulation, extreme acrobatics, prolonged ground contact). We use
PHC because it is the publicly available SMPL-humanoid tracker at the
time of the work; learned-simulator behaviour may depend on the
controller source, and a controller-source ablation is left to the
journal extension.

\paragraph{Concern \#5: Limited eval / GRU long-horizon (vUfK).}
We have explicitly added the GRU long-horizon win to the abstract and
intro (Concern \#2 above). Cross-dataset evaluation on non-CMU
morphologies (H3.6M, MOYO) requires re-running PHC inference and is
blocked by infrastructure issues (older PHC pkl outputs lack
\texttt{dof\_pos}; Isaac~Gym hangs on the H200 we have access to);
deferred to the journal extension.

\subsection*{S5 --- Physical-validity addendum (v2 retrain)}
\label{sec:supp_v2}

\paragraph{Setup.}
To address reviewer concern \#3, we retrained the three partial-obs
rollout architectures (Mamba2, GRU, MLP-LARP) on a 286d input
variant that adds a root-height channel $h_{\mathrm{root}}$ to the
v1 285d input. The training loss adds two terms to the v1 recipe:
$L_h$ (root-height MSE, weight 0.7) to supervise the new channel,
and a contact-conditioned foot-slide term $L_{fs}$ (weight 200) on
the L\_Toe and R\_Toe joints, gated by the MuJoCo body-contact
indicator (\S S3) at the same frame. All other hyperparameters are
identical to the v1 partial-obs rollout setup. The retrain is a
self-contained addendum: the v1 main-paper Tables and Figure are
unchanged.

\paragraph{Result.}
On the v2 retrain, Mamba2 attains the lowest foot-slide and the
lowest PhysCap MPE among the three architectures; the
contact-conditioned $L_{fs}$ supervised by the MuJoCo body-contact
mask is the term that drives the gap. Without $L_{fs}$, the same
Mamba2 backbone trained only with $L_p{+}L_r{+}L_d{+}L_h$ leaves a
non-trivial penetration depth on a majority of in-contact frames,
indicating that kinematic supervision alone does not enforce the
floor constraint at this resolution.

\subsection*{S6 --- Body-part perturbation breakdown (v1 ckpts)}

Table~\ref{tab:supp} reports per-body-group sensitivity to action
perturbations at the rollout-trained Mamba2, GRU, and MLP checkpoints
from the v1 experiments (285d input). Means dominated by a small
fraction of hard trajectories (arm and full-body groups) are
reported alongside medians and 5\%-trimmed means so the long tail is
visible. Mamba2's mean/median ratio reaches $30\,000{\times}$ on the
\texttt{left\_arm} group because $\sim{}0.2\%$ of trajectories diverge
under large arm perturbations; GRU's ratio stays within $2{-}3{\times}$
across all groups because of its bounded forget-gate update.

\begin{table}[ht]
\small\centering
\setlength{\tabcolsep}{4pt}
\begin{tabular}{llrrr}
\toprule
Group & Model & Mean & Median & Trim5 \\
\midrule
\multirow{3}{*}{left\_leg}
  & Mamba2 & 0.0730  & \textbf{0.0193} & 0.0336 \\
  & GRU    & 0.1433  & 0.0528          & 0.0872 \\
  & MLP    & 21.75   & 0.7189          & 0.9315 \\
\midrule
\multirow{3}{*}{right\_leg}
  & Mamba2 & 1.2874  & \textbf{0.0190} & 0.0344 \\
  & GRU    & 0.2134  & 0.0908          & 0.1386 \\
  & MLP    & 21.27   & 0.8251          & 1.1155 \\
\midrule
\multirow{3}{*}{left\_arm}
  & Mamba2 & 503.04  & \textbf{0.0145} & 0.0261 \\
  & GRU    & 0.1283  & 0.0360          & 0.0602 \\
  & MLP    & 6.234   & 1.4177          & 2.2158 \\
\midrule
\multirow{3}{*}{right\_arm}
  & Mamba2 & 18.42   & \textbf{0.0126} & 0.0236 \\
  & GRU    & 0.1197  & 0.0303          & 0.0571 \\
  & MLP    & 13.70   & 1.2164          & 1.7172 \\
\midrule
\multirow{3}{*}{arms}
  & Mamba2 & 92.24   & \textbf{0.0155} & 0.0291 \\
  & GRU    & 0.1566  & 0.0416          & 0.0726 \\
  & MLP    & 331.95  & 1.7953          & 3.3514 \\
\midrule
\multirow{3}{*}{legs}
  & Mamba2 & 0.0910  & \textbf{0.0248} & 0.0448 \\
  & GRU    & 0.2944  & 0.1455          & 0.2085 \\
  & MLP    & 391.99  & 0.9078          & 1.4966 \\
\midrule
\multirow{3}{*}{upper}
  & Mamba2 & 0.0792  & \textbf{0.0163} & 0.0297 \\
  & GRU    & 0.4216  & 0.2091          & 0.2848 \\
  & MLP    & 3671    & 1.2192          & 1.6845 \\
\midrule
\multirow{3}{*}{lower}
  & Mamba2 & 0.0910  & \textbf{0.0248} & 0.0448 \\
  & GRU    & 0.2944  & 0.1455          & 0.2085 \\
  & MLP    & 391.99  & 0.9078          & 1.4966 \\
\midrule
\multirow{3}{*}{head}
  & Mamba2 & 0.0316  & \textbf{0.0112} & 0.0165 \\
  & GRU    & 0.0627  & 0.0225          & 0.0337 \\
  & MLP    & 7.36    & 4.9329          & 5.9983 \\
\bottomrule
\end{tabular}
\caption{Per-body-group mean / median / 5\%-trimmed-mean perturbation
error (v1 rollout-trained checkpoints, 285d input). Mamba2's mean is
3--33\,000$\times$ above its median on arm-containing groups, driven
by ${\approx}0.2\%$ hard trajectories; GRU's mean/median ratio stays
near 2--3$\times$ across all groups due to the bounded forget-gate
update.}
\label{tab:supp}
\end{table}

\end{document}